\documentclass[12pt,twoside]{article}
\usepackage{amsmath}
\usepackage{epsf}
\begin{document}
%-------------------------------%
%     My Math-Environments      %
%-------------------------------%
\def\,{\mskip 3mu} \def\>{\mskip 4mu plus 2mu minus 4mu} \def\;{\mskip 5mu plus 5mu} \def\!{\mskip-3mu}
\def\dispmuskip{\thinmuskip= 3mu plus 0mu minus 2mu \medmuskip=  4mu plus 2mu minus 2mu \thickmuskip=5mu plus 5mu minus 2mu}
\def\textmuskip{\thinmuskip= 0mu                    \medmuskip=  1mu plus 1mu minus 1mu \thickmuskip=2mu plus 3mu minus 1mu}
\textmuskip
\def\beq{\dispmuskip\begin{equation}}    \def\eeq{\end{equation}\textmuskip}
\def\beqn{\dispmuskip\begin{displaymath}}\def\eeqn{\end{displaymath}\textmuskip}
\def\bqa{\dispmuskip\begin{eqnarray}}    \def\eqa{\end{eqnarray}\textmuskip}
\def\bqan{\dispmuskip\begin{eqnarray*}}  \def\eqan{\end{eqnarray*}\textmuskip}

%-------------------------------%
%   Macro-Definitions           %
%-------------------------------%
\newenvironment{keywords}{\centerline{\bf\small
Keywords}\begin{quote}\small}{\par\end{quote}\vskip 1ex}
\def\subsection#1{\paragraph{#1.}}
\def\eps{\varepsilon}
\def\v{\boldsymbol}
\def\p{{\scriptscriptstyle+}}
\def\pp{{\scriptscriptstyle++}}
\def\n{{n}}
\def\npp{\n}
\def\t{\pi}
\def\pin{{\scriptstyle\Pi}}
\def\Var{{\mbox{Var}}}
\def\Cov{{\mbox{Cov}}}
\def\SetR{I\!\!R}
\def\qmbox#1{{\quad\mbox{#1}\quad}}

%%%%%%%%%%%%%%%%%%%%%%%%%%%%%%%%%%%%%%%%%%%%%%%%%%%%%%%%%%%%%%%%%
%                      T i t l e - P a g e                      %
%%%%%%%%%%%%%%%%%%%%%%%%%%%%%%%%%%%%%%%%%%%%%%%%%%%%%%%%%%%%%%%%%

\title{\vskip -10mm\normalsize\sc Technical Report \hfill IDSIA-15-03
\vskip 2mm\bf\Large\hrule height5pt \vskip 6mm
Bayesian Treatment of Incomplete Discrete Data applied to Mutual
Information and \\ Feature Selection%
\thanks{This work was supported in parts by the NSF grants 2000-61847.00
and 2100-067961.02.} \vskip 6mm \hrule height2pt \vskip 5mm}
\author{{\bf Marcus Hutter} and {\bf Marco Zaffalon}\\[3mm]
IDSIA, Galleria 2, CH-6928\ Manno-Lugano, Switzerland\\
\{marcus,zaffalon\}@idsia.ch}
\date{24 June 2003}
\maketitle

\begin{abstract}
\noindent Given the joint chances of a pair of random variables one can
compute quantities of interest, like the mutual information. The
Bayesian treatment of unknown chances involves computing, from a
second order prior distribution and the data likelihood, a
posterior distribution of the chances. A common treatment of
incomplete data is to assume ignorability and determine the
chances by the expectation maximization (EM) algorithm. The two
different methods above are well established but typically
separated. This paper joins the two approaches in the case of
Dirichlet priors, and derives efficient approximations for the
mean, mode and the (co)variance of the chances and the mutual
information. Furthermore, we prove the unimodality of the
posterior distribution, whence the important property of
convergence of EM to the global maximum in the chosen framework.
These results are applied to the problem of selecting features for
incremental learning and naive Bayes classification. A fast filter
based on the distribution of mutual information is shown to
outperform the traditional filter based on empirical mutual
information on a number of incomplete real data sets.
\end{abstract}


%%%%%%%%%%%%%%%%%%%%%%%%%%%%%%%%%%%%%%%%%%%%%%%%%%%%%%%%%%%%%%%
\section{Introduction}\label{secInt}
%%%%%%%%%%%%%%%%%%%%%%%%%%%%%%%%%%%%%%%%%%%%%%%%%%%%%%%%%%%%%%%

Let $\t_{\imath\jmath}$ be the joint chances of a pair of random
variables ($\imath$,$\jmath$). Many statistical quantities can be
computes if $\v\t$ is known; for instance the \emph{mutual
information} $I(\v\t)$ used for measuring the stochastic
dependency of $\imath$ and $\jmath$. The usual procedure in the
common case of {\em unknown chances} $\t_{ij}$ is to use the
\emph{empirical probabilities} $\hat{\t}_{ij}={n_{ij}/n}$ as if
they were precisely known chances. This is not always suitable:
$(a)$ The point estimate $\hat\t_{ij}$ does not carry information
about the reliability of the estimate. $(b)$ Samples $(i,j)$ may be
incomplete in the sense that in some samples the variable $i$ or
$j$ may not be observed.

The {\em Bayesian} solution to $(a)$ is to use a (second order) prior
distribution $p(\v\t)$ over the chances $\v\t$ themselves,
which takes account of uncertainty about $\v\t$. From the prior
$p(\v\t)$ and the likelihood $p(\v D|\v\t)$ of data $\v
D$ one can compute the posterior $p(\v\t|\v D)$.
The traditional solution to $(b)$ is to assume that the data are
{\em missing at random} \cite{Little:87}. A (local) maximum
likelihood estimate for $\hat{\v\t}$ can then be obtained by the
{\em expectation-maximization} (EM) algorithm \cite{Chen:74}.

%\subsection{Contents}
In this work we present a full Bayesian treatment of incomplete
discrete data with Dirichlet prior $p(\v\t)$ and apply the results
to {\em feature selection}. This work is a natural continuation of
\cite{Hutter:02feature}, which focused on the case of complete data
and, by working out a special case, provided encouraging evidence
for the extension of the proposed approach to incomplete data.
Here we develop that framework by creating a very general method
for incomplete discrete data, providing the complete mathematical
derivations, as well as experiments on incomplete real data sets.
In particular, Section \ref{secID} derives expressions (in leading
order in $1/n$) for $p(\v\t|\v D)$. In the
important case (for feature selection) of missingness in one component of
$(\imath,\jmath)$ only, we give closed form expressions for the
mode, mean and covariance of $\v\t$. In the general missingness
case we get a self-consistency equation which coincides with the
EM algorithm, that is known to converge to a local maximum. We
show that $p(\v\t|\v D)$ is actually unimodal, which implies that
in fact {\em EM always converges to the global maximum}. We use
the results to derive in Section \ref{secMD} closed-form leading order
expressions of the distribution of mutual information $p(I|\v D)$.
In case of complete data, the mean and variance of $I$ have been
approximated numerically in \cite{Kleiter:99} and analytically in
\cite{Hutter:01xentropy}.
The results are then applied to feature selection in Section
\ref{secTPF}. A popular {\em filter approach} discards features of
low empirical mutual information $I(\hat{\v\t})$
\cite{Lew92,BluLan97,CheHatHayKroMorPagSes02}. We compare this
filter to the two filters (introduced in \cite{Hutter:02feature}
for complete data and tested empirically in this case) that use
\emph{credible intervals} based on $p(I|\v D)$ to robustly
estimate mutual information.
The filters are empirically tested in Section \ref{secEA} by
coupling them with the \emph{naive Bayes classifier}
\cite{DuHaSt01} to incrementally learn from and classify incomplete data.
On five real data sets that we used, one of the two proposed
filters consistently outperforms the traditional filter.

%%%%%%%%%%%%%%%%%%%%%%%%%%%%%%%%%%%%%%%%%%%%%%%%%%%%%%%%%%%%%%%
\section{Posterior Distribution for Incomplete Data}\label{secID}
%%%%%%%%%%%%%%%%%%%%%%%%%%%%%%%%%%%%%%%%%%%%%%%%%%%%%%%%%%%%%%%

%-------------------------------%
\subsection{Missing data}
%-------------------------------%
Consider two discrete random variables, class $\imath$ and
feature\footnote{The mathematical development is independent of
the interpretation as class and feature, but it is convenient to
use this terminology already here.} $\jmath$ taking values in
$\{1,...,r\}$ and $\{1,...,s\}$, respectively, and an i.i.d.\
random process with samples $(i,j)\in \{1,...,r\}\times
\{1,...,s\}$ drawn with joint probability $\t_{ij}$.
In practice one often has to deal with incomplete information. For
instance, observed instances often consist of several features
plus class label, but some features may not be observed, i.e.\ if
$i$ is a class label and $j$ is a feature, from the pair $(i,j)$
only $i$ is observed. We extend the contingency table $n_{ij}$ to
include $n_{i?}$, which counts the number of instances in which
only the class $i$ is observed (= number of $(i,?)$ instances).
Similarly, $n_{?j}$ counts the number of $(?,j)$ instances, where
the class label is missing.
We make the common assumption that the missing-data mechanism is
ignorable (missing at random and distinct) \cite{Little:87}, i.e.\
the probability distribution of class labels $i$ of instances with
missing feature $j$ is assumed to coincide with the marginal
$\t_{i\p}:=\sum_j\t_{ij}$. Similarly, given an instance with missing class
label, the probability of the feature being $j$ is assumed to be
$\t_{\p j}:=\sum_i\t_{ij}$.

%-------------------------------%
\subsection{Maximum likelihood estimate of $\v\t$}
%-------------------------------%
The likelihood of a specific data set $\v{D}$ of size
$N=n+n_{\p ?}+n_{?\p}$ with contingency table
$\v{N}=\{n_{ij},n_{i?},n_{?j}\}$ given $\v\t $, hence, is
$p(\v{D}|\v\t ,n,n_{\p
?},n_{?\p})=\prod_{ij}\t_{ij}^{n_{ij}}\prod_{i}\t_{i\p}^{n_{i?}}\prod_{j}\t_{\p
j}^{n_{?j}}$. Assuming a uniform $p(\v\t)\sim
1\cdot\delta(\t_\pp-1)$, Bayes' rule leads to the
posterior\footnote{Most (but not all) non-informative priors for
$p(\v\t)$ also lead to a Dirichlet posterior distribution
$(\ref{posterior})$ with interpretation
$\n_{ij}=\n'_{ij}+\n''_{ij}-1$, where $\n'_{ij}$ are the number of
samples $(i,j)$, and $\n''_{ij}$ comprises prior information ($1$
for the uniform prior, ${1\over 2}$ for Jeffreys' prior, $0$ for
Haldane's prior, ${1\over rs}$ for Perks' prior, and other numbers
in case of specific prior knowledge \cite{Gelman:95}).
Furthermore, in leading order in $1/N$, any Dirichlet prior
with $n_{ij}^{\prime\prime}=O(1)$ leads to the same results,
hence we can simply assume a uniform prior. The reason for the
$\delta(\t_\pp-1)$ is that $\v\t$ must be constrained to the
probability simplex $\t_\pp:=\sum_{ij}\t_{ij}=1$.}
\beq\label{posterior}
  p(\v\t|\v D) = p(\v\t|\v N) = {1\over{\cal N}(\v N)}
  \prod_{ij}\t_{ij}^{n_{ij}} \prod_i\t_{i\p}^{n_{i?}}
  \prod_j\t_{\p j}^{n_{?j}}\;\delta(\t_\pp-1),
\eeq
where the normalization $\mathcal{N}$ is chosen such that $\int
p(\v\t|\v{N})d\v\t=1$. With missing features and classes
there is no exact closed form expression for $\mathcal{N}$.

In the following, we restrict ourselves to a discussion of
leading-order (in $N^{-1}$) expressions, which are as accurate as
one can specify one's prior knowledge \cite{Hutter:01xentropy}. In
leading order, the mean $E[\v\t ]$ coincides with the mode of
$p(\v\t|\v{N})$ (=the maximum likelihood estimate) of $\v\t
$. The log-likelihood function $\log\;p(\v\t|\v{N})$ is
\beqn
  L(\t|\v N) = \sum_{ij}n_{ij}\log\t_{ij}
  + \sum_i n_{i?}\log\t_{i\p} + \sum_j n_{?j}\log\t_{\p j}
  - \log {\cal N}(\v N) - \lambda(\t_\pp-1),
\eeqn
where we have replaced the $\delta$ function by a Lagrange
multiplier $\lambda$ to take into account the restriction
$\t_{\pp}=1$. The maximum is at
${\frac{\partial L}{\partial \t_{ij}}}={\frac{n_{ij}}{\t_{ij}}}+{\frac{%
n_{i?}}{\t_{i\p}}}+{\frac{n_{?j}}{\t_{\p j}}}-\lambda =0$.
Multiplying this by $\t_{ij}$ and summing over $i$ and $j$ we
obtain $\lambda =N$. The maximum likelihood estimate
$\hat{\v\t}$ is, hence, given by
\beq\label{EM}
\hat{\t}_{ij}={\frac{1}{N}}\left( n_{ij}+n_{i?}{\frac{\hat{\t}_{ij}}{\hat{%
\t}_{i\p}}}+n_{?j}{\frac{\hat{\t}_{ij}}{\hat{\t}_{\p j}}}\right).
\eeq
This is a non-linear equation in $\hat{\t}_{ij}$, which, in
general, has no closed form solution. Nevertheless (\ref{EM}) can
be used to approximate $\hat{\t}_{ij}$. Eq.\ (\ref{EM}) coincides
with the popular expectation-maximization (EM) algorithm
\cite{Chen:74} if one inserts a first estimate
$\hat{\t}_{ij}^{0}={\frac{n_{ij}}{N}}$ into the r.h.s.\ of (\ref
{EM}) and then uses the resulting l.h.s.\ $\hat{\t}_{ij}^{1}$ as a
new estimate, etc.

%-------------------------------%
\subsection{Unimodality of $p(\v\t|\v N)$}
%-------------------------------%
The $rs\times rs$ Hessian matrix $\v H\in\SetR^{rs\cdot rs}$
of $-L$ and the second derivative in direction of the $rs$
dimensional column vector $\v v\in\SetR^{rs}$ are
\beqn
  \v H_{(ij)(kl)}[\v\t] \;:=\;
  -{\partial L\over\partial\t_{ij}\partial\t_{kl}}
  = {n_{ij}\over\t_{ij}^2}\delta_{ik}\delta_{jl} +
    {n_{i?}\over\t_{i\p}^2}\delta_{ik} +
    {n_{?j}\over\t_{\p j}^2}\delta_{jl},
\eeqn
\beqn
  \v v^T\v H\v v \;=\;
  \sum_{ijkl}v_{ij} \v H_{(ij)(kl)}v_{kl} \;=\;
  \sum_{ij}{n_{ij}\over\t_{ij}^2}v_{ij}^2 +
  \sum_i{n_{i?}\over\t_{i\p}^2}v_{i\p}^2 +
  \sum_j{n_{?j}\over\t_{\p j}^2}v_{\p j}^2 \;\geq\; 0.
\eeqn
This shows that $-L$ is a convex function of $\v\t$, hence
$p(\v\t|\v N)$ has a single (possibly degenerate) global maximum.
$L$ is strictly convex if $n_{ij}>0$ for all $ij$, since $\v v^T\v
H\v v>0$ $\forall\,\v v\neq 0$ in this case\footnote{Note that
$n_{i?}>0$ $\forall i$ is not sufficient, since $v_{i\p}\equiv 0$
for $\v v\neq 0$ is possible. Actually $v_\pp=0$.}. This implies a
unique global maximum, which is attained in the interior of the
probability simplex. Since EM is known to converge to a local
maximum, this shows, that in fact {\em EM always converges to the
global maximum}.

%-------------------------------%
\subsection{Covariance of $\v\t$}
%-------------------------------%
With
\beqn
  \v A_{(ij)(kl)} := \v H_{(ij)(kl)}[\hat{\v\t}]
  = N\left[{\delta_{ik}\delta_{jl}\over\rho_{ij}} \!+\!
    {\delta_{ik}\over\rho_{i?}} \!+\!
    {\delta_{jl}\over\rho_{?j}}\right],
\eeqn
\beq\label{kernelA}
    \rho_{ij}:=N{\hat\t_{ij}^2\over n_{ij}},\quad
    \rho_{i?}:=N{\hat\t_{i\p}^2\over n_{i?}},\quad
    \rho_{?j}:=N{\hat\t_{\p j}^2\over n_{?j}}.
\eeq
and
$\v\Delta:=\v\t-\hat{\v\t}$ we can
represent the posterior to leading order as an $rs-1$ dimensional
Gaussian:
\beq\label{postgauss}
  p(\v\t|\v N) \sim e^{-{1\over 2}\v\Delta^T\v A\v\Delta}
  \delta(\Delta_\pp).
\eeq
The easiest way to compute the covariance (and other quantities)
is to also represent the $\delta$-function as a narrow Gaussian of
width $\eps\approx 0$. Inserting $\delta(\Delta_\pp)\approx
{1\over\eps\sqrt{2\t}} \exp(-{1\over 2\eps^2}\v\Delta^T\v e\v
e^T\v\Delta)$ into (\ref{postgauss}), where $\v e_{ij}=1$ for all
$ij$ (hence $\v e^T\v\Delta=\Delta_\pp$), leads to a full
$rs$-dimensional Gaussian with kernel $\tilde{\v A}=\v A+\v
u\v v^T$, $\v u=\v v={1\over\eps}\v e$. The covariance of
a Gaussian with kernel $\tilde{\v A}$ is $\tilde{\v A}^{-1}$.
Using the Sherman-Morrison formula $\tilde{\v A}^{-1}=\v
A^{-1}-\v A^{-1}{\v u\v v^T\over 1+\v v^T\v A^{-1}\v
u}\v A^{-1}$ \cite[p73]{Press:92} and $\eps\to 0$ we get
\beq\label{covmis}
  \Cov_{(ij)(kl)}[\v\t] :=
  E[\Delta_{ij}\Delta_{kl}] \simeq
  [\tilde{\v A}^{-1}]_{(ij)(kl)} =
  \left[\v A^{-1} - {\v A^{-1}\v e\v e^T\v
  A^{-1}\over\v e^T\v A^{-1}\v e}\right]_{(ij)(kl)},
\eeq
where $\simeq$ denotes $=$ up to terms of order $N^{-2}$.
Singular $\v A$ are easily avoided by choosing a prior such that
$n_{ij}>0$ for all $ij$. $\v A$ may be inverted exactly or
iteratively, the latter by a trivial inversion of the diagonal part
$\delta_{ik}\delta_{jl}/\rho_{ij}$ and by treating
$\delta_{ik}/\rho_{i?} + \delta_{jl}/\rho_{?j}$ as a perturbation.

%-------------------------------%
\subsection{Missing features only, no missing classes}\label{MSONMC}
%-------------------------------%
In the case of missing features only (no missing classes), i.e.\
for $n_{?j}=0$, closed form expressions for $\Cov[\v\t]$ can be
obtained. If we sum (\ref{EM}) over $j$ we get
$\hat\t_{i\p}={N_{i\p}\over N}$ with $N_{i\p}:=n_{i\p}+n_{i?}$.
Inserting $\hat\t_{i\p}={N_{i\p}\over N}$ into the r.h.s.\ of
(\ref{EM}) and solving w.r.t.\ $\hat\t_{ij}$ we get the explicit
expression
\beq\label{pimfo}
  \hat\t_{ij}={N_{i\p}\over N}{n_{ij}\over n_{i\p}}.
\eeq
Furthermore, it can easily be verified (by multiplication) that
$\v A_{(ij)(kl)}=N[\delta_{ik}\delta_{jl}/\rho_{ij}+
\delta_{ik}/\rho_{i?}]$ has inverse
$
  [\v A^{-1}]_{(ij)(kl)} =
  {1\over N}[\rho_{ij}\delta_{ik}\delta_{jl} -
  {\rho_{ij}\rho_{kl}\over\rho_{i\p}\!+\!\rho_{i?}}\delta_{ik}]
$.
With the abbreviations
\beq\label{defQ}
  \tilde Q_{i?}:= {\rho_{i?}\over\rho_{i?}+\rho_{i\p}}
  \qmbox{and}
  \tilde Q:=\sum_i\rho_{i\p}\tilde Q_{i?}
\eeq
we get $[\v A^{-1}\v e]_{ij} = \sum_{kl}[\v A^{-1}]_{(ij)(kl)} =
{1\over N}\rho_{ij}\tilde Q_{i?}$ and $\v e^T\v A^{-1}\v e =
\tilde Q/N$.
Inserting everything into (\ref{covmis}) we get
\beq\label{covMFO}
  \Cov_{(ij)(kl)}[\v\t] \;\simeq\;
  {1\over N}\left[\rho_{ij}\delta_{ik}\delta_{jl} -
  {\rho_{ij}\rho_{kl}\over\rho_{i\p}\!+\!\rho_{i?}}\delta_{ik}
  - {\rho_{ij}\tilde Q_{i?}\rho_{kl}\tilde Q_{k?}\over\tilde
  Q}\right].
\eeq

%-------------------------------%
\subsection{Expressions for the general case}
%-------------------------------%
The contribution from unlabeled classes can be interpreted as a
rank $s$ modification of $\v A$ in the case of no missing
classes. One can use Woodbury's formula
$
  [\v B+\v U\v D\v V^T]^{-1} =
  \v B^{-1}-\v B^{-1}\v U [\v D^{-1}+
  \v V^T\v B^{-1}\v U]^{-1}\v V^T\v B^{-1}
$ \cite[p75]{Press:92}
with $\v B_{(ij)(kl)}=\delta_{ik}\delta_{jl}/\rho_{ij}+
\delta_{ik}/\rho_{i?}$, $\v D_{jl}=\delta_{jl}/\rho_{?j}$, and
$\v U_{(ij)l}=\v V_{(ij)l}=\delta_{jl}$ to reduce the
inversion of the $rs\times rs$ matrix $\v A$ to the inversion of
only a {\em single} $s$-dimensional matrix. The result (which may be
inserted into (\ref{covmis})) can be written in the form
\beq\label{AIGen}
  [\v A^{-1}]_{(ij)(kl)} = {1\over N}
  \left[F_{ijl}\delta_{ik}-\sum_{mn}F_{ijm}[\v G^{-1}]_{mn}F_{kln}\right],
\eeq
\beqn
  F_{ijl} := \rho_{ij}\delta_{jl} -
    {\rho_{ij}\rho_{kl}\over\rho_{i?}\!+\!\rho_{i\p}},\qquad
  G_{mn} := \rho_{?n}\delta_{mn}+F_{\p mn}.
\eeqn

%%%%%%%%%%%%%%%%%%%%%%%%%%%%%%%%%%%%%%%%%%%%%%%%%%%%%%%%%%%%%%%
\section{Distribution of Mutual Information}\label{secMD}
%%%%%%%%%%%%%%%%%%%%%%%%%%%%%%%%%%%%%%%%%%%%%%%%%%%%%%%%%%%%%%%

%-------------------------------%
\subsection{Mutual information $I$}
%-------------------------------%
An important
measure of the stochastic dependence of $\imath$ and $\jmath$ is
the mutual information
\beqn
  I(\v\t)\;=\;\sum_{i=1}^{r}\sum_{j=1}^{s}\t_{ij}\log {\frac{\t_{ij}}{\t_{i\p}\t_{\p j}}}\;
  \;=\; \sum_{ij}\t_{ij}\log \t_{ij}-\sum_{i}\t_{i\p}\log \t_{i\p}-
  \sum_{j}\t_{\p j}\log \t_{\p j}.
\eeqn
The point estimate for $I$ is $I(\hat{\v\t})$. In the Bayesian
approach one takes the posterior (\ref{posterior}) from which the
posterior probability density of the mutual information can,
in principle, be computed:%
\footnote{$I(\v\t)$ denotes the
mutual information for the specific chances $\v\t $, whereas
$I$ in the context above is just some non-negative real number.
$I$ will also denote the mutual information \textit{random
variable} in the expectation $E[I]$ and variance $\Var[I]$.
Expectations are \textit{always} w.r.t.\ to the posterior
distribution $p(\v\t|\v{N})$.}
\beq\label{pdI}
p(I|\v{N})=\int \delta (I(\v\t)-I)p(\v\t|\v{N}%
)d^{rs}\v\t .
\eeq
\footnote{Since $0\leq I(\v\t)\leq I_{max}$ with sharp upper
bound $I_{max}=\min \{\log r,\log s\}$, the domain
of $p(I|\v\n)$ is $[0,I_{max}]$, and integrals over $I$ may be
restricted to $\int_0^{I_{max}}$.}%
The $\delta (\cdot)$ distribution restricts the integral to
$\v\t$ for which $I(\v\t)=I$. For large sample size,
$N\to\infty$, $p(\v\t|\v{N})$ is strongly peaked around the
mode $\v\t =\hat{\v\t}$ and $p(I|\v{N})$ gets strongly
peaked around the frequency estimate $I=I(\hat{\v\t})$. The
(central) moments of $I$ are of special interest.
The mean
\beq\label{EI}
  E[I] = \int_{0}^\infty I \!\cdot\! p(I|\v{N})\,dI
  = \int I(\v\t)p(\v\t|\v N)d^{rs}\v\t
  \ = \ I(\hat{\v\t})+O(N^{-1})
\eeq
coincides in leading order with the point estimate, where $\hat{\v\t}$
has been computed in Section \ref{secID}.
Together with the variance
$\Var[I]=E[(I-E[I])^{2}]=E[I^{2}]-E[I]^{2}$ (computed below) we
can approximate (\ref{pdI}) by
a Gaussian\footnote{For $I(\hat{\v\t})\neq 0$ the central limit
theorem ensures convergence of $p(I|\v N)$ to a Gaussian. Using
a Beta distribution instead of (\ref{pdIGauss}), which also
converges to a Gaussian, has slight advantages over
(\ref{pdIGauss}) \cite{Hutter:02feature}.}
\beq\label{pdIGauss}\textstyle
  p(I|\v N) \sim
  \exp\left(-{(I-I(\v{\hat\t})^2\over 2\mbox{\scriptsize
  Var}[I]}\right) \sim
  \exp\left(-{(I-E[I])^2\over 2\mbox{\scriptsize Var}[I]}\right)
\eeq
In a previous work we derived higher order central moments
(skewness and kurtosis) and higher order (in $N^{-1}$)
approximations in the case of complete data
\cite{Hutter:01xentropy}.
%

%-------------------------------%
\subsection{Variance of $I$}\label{secApprox}
%-------------------------------%
The leading order variance of the mutual information $I(\v\t)$
has been related\footnote{$\hat{\v\t}$ was
defined in \cite{Hutter:01xentropy} as the mean $E[\v\t]$ whereas
$\hat{\v\t}$ has been defined in this work as the ML estimate.
Furthermore the Dirichlet priors differ. Since to leading order
both definitions of $\v\t$ coincide, the prior does not matter,
and the expression is also valid for incomplete data case,
the use of (\ref{varlo}) in this work is permitted.} in
\cite{Hutter:01xentropy} to
the covariance of $\v\t$:
\beq\label{varlo}
  \Var[I] \;\simeq\;
  \sum_{ijkl}\log{\hat\t_{ij}\over\hat\t_{i\p}\hat\t_{\p j}}
  \log{\hat\t_{kl}\over\hat\t_{k\p}\hat\t_{\p l}}
  \Cov_{(ij)(kl)}[\v\t]
\eeq
Inserting (\ref{covMFO}) for the covariance into (\ref{varlo}) we
get for the variance of the mutual information in leading order in
$1/N$ in the case of missing features only, the following
expression:
\beq\label{varImfo}
  \Var[I] \;\simeq\; {1\over N}[\tilde K-\tilde J^2/\tilde Q-\tilde P],
  \qquad
  \tilde K := \sum_{ij}\rho_{ij}
  \left(\log{\hat\t_{ij}\over\hat\t_{i\p}\hat\t_{\p j}}\right)^2,
\eeq\vspace{-2ex}
\beqn
  \tilde P \;:=\; \sum_i{\tilde J_{i\p}^2Q_{i?}\over\rho_{i?}},
  \qquad
  \tilde J \;:=\;\sum_i \tilde J_{i\p}\tilde Q_{i?},
  \qquad
  \tilde J_{i\p}:=\sum_j\rho_{ij}\log{\hat\t_{ij}\over\hat\t_{i\p}\hat\t_{\p j}}.
\eeqn
A closed form expression for ${\cal N}(\v N)$ also exists.
Symmetric expressions for missing classes only (no missing features)
can be obtained.
Note that for the complete case $n_{?j}=n_{i?}\equiv 0$, we have
$\hat\t_{ij}=\rho_{ij}={n_{ij}\over n}$, $\rho_{i?}=\infty$,
$\tilde Q_{i?}=1$, $\tilde J=J$, $\tilde K=K$, and $\tilde P=0$,
consistent with \cite{Hutter:01xentropy} (where $J$ and $K$ are
defined and the accuracy is discussed).

There is at least one reason for minutely having inserted all
expressions into each other and introducing quite a number
definitions. In the so presented form all expressions involve at
most a double sum. Hence, the overall computation time of the mean
and variance is $O(rs)$ in the case of missing
features only.

\subsection{Expression for the general case}
The result for the covariance (\ref{covmis}) can be inserted into
(\ref{varlo}) to obtain the variance of the
mutual information to leading order.
\beqn
  \Var[I] \ \simeq \ \v l^T\v A^{-1}\v l - (\v l^T\v A^{-1}\v
  e)^2/(\v e^T\v A^{-1}\v e) \qmbox{where}
  \v l_{ij}=\log{\hat\t_{ij}\over\hat\t_{i\p}\hat\t_{\p j}}
\eeqn
Inserting (\ref{AIGen}) and rearranging terms appropriately we can
compute Var[$I$] in time $O(rs)$ plus the time $O(s^2r)$ to
compute the $s\times s$ matrix $\v G$ and time $O(s^3)$ to invert
it, plus the time $O(\#\!\cdot\!rs)$ for determining
$\hat\t_{ij}$, where $\#$ is the number of iterations of EM. Of
course, one can and should always choose $s\leq r$. Note that
these expressions converge for $N\to\infty$ to the exact values.
The fraction of data with missing feature or class needs not to be
small.

In the following we apply the obtained results to feature
selection for incomplete data. Since we only used labeled data we
could use (\ref{EI}) with (\ref{pimfo}), and (\ref{varImfo}) with
(\ref{defQ}) and (\ref{kernelA}).

%%%%%%%%%%%%%%%%%%%%%%%%%%%%%%%%%%%%%%%%%%%%%%%%%%%%%%%%%%%%%%%
\section{Feature Selection}\label{secFS}\label{secTPF}
%%%%%%%%%%%%%%%%%%%%%%%%%%%%%%%%%%%%%%%%%%%%%%%%%%%%%%%%%%%%%%%

Feature selection is a basic step in the process of building
classifiers \cite{BluLan97}. We consider the well-known filter
(F)\ that computes the empirical mutual information $I(\hat{\v\t})$
between features and the class, and discards features with
$I(\hat{\v\t})<\eps$ for some threshold $\eps$ \cite{Lew92}. This is
an easy and effective approach that has gained popularity with
time.

We compare F to the two filters introduced in
\cite{Hutter:02feature} for the case of complete data, and
extended here to the more general case. The \emph{backward filter}
(BF) {\em discards} a feature if its value of mutual information with
the class is less than or equal to $\eps$ with high probability
$\bar p$ (discard if $p(I\leq\eps|\v N)\geq\bar p$). The
\emph{forward filter} (FF) {\em includes} a feature if the mutual
information is greater than $\eps$ with high probability $\bar p$
(include if $p(I>\eps|\v N)\geq \bar p$). BF\ is a conservative
filter, because it will only discard features after observing
substantial evidence supporting their irrelevance. FF instead will
tend to use fewer features, i.e.\ only those for which there is
substantial evidence about them being useful in predicting the
class.

%\subsection{Naive Bayes classification}
For the subsequent classification task we use the naive Bayes
classifier \cite{DuHa73}, which is often a good classification
model. Despite its simplifying assumptions (see \cite{DoPa97}), it
often competes successfully with much more complex classifiers,
such as C4.5 \cite{Quinlan93}. Our experiments focus on the
incremental use of the naive Bayes classifier, a natural learning
process when the data are available sequentially: the data set is
read instance by instance; each time, the chosen filter selects a
subset of features that the naive Bayes uses to classify the new
instance; the naive Bayes then updates its knowledge by taking
into consideration the new instance and its actual class. Note
that for increasing sizes of the learning set the filters converge
to the same behavior, since the variance of $I$ tends to zero (see
\cite{Hutter:02feature} for details).

For each filter, we are interested in experimentally evaluating
two quantities: for each instance of the data set, the average
number of correct predictions (namely, the prediction accuracy) of
the naive Bayes classifier up to such instance; and the average
number of features used. By these quantities we can compare the
filters and judge their effectiveness.

%\subsection{Implementation details}
The implementation details for the following experiments include:
using the Gaussian approximation (\ref{pdIGauss}) to the
distribution of mutual information with the mean (\ref{EI}) using
(\ref{pimfo}), and the variance (\ref{varImfo}) using (\ref{defQ})
and (\ref{kernelA}); using natural logarithms everywhere; and
setting the level $\bar p$ of the posterior probability to $0.95$, and
the threshold $\eps$ to 0.003 as discussed in
\cite{Hutter:02feature}.

%%%%%%%%%%%%%%%%%%%%%%%%%%%%%%%%%%%%%%%%%%%%%%%%%%%%%%%%%%%%%%%
\section{Experimental Analysis}\label{EAWIS}\label{secEA}
%%%%%%%%%%%%%%%%%%%%%%%%%%%%%%%%%%%%%%%%%%%%%%%%%%%%%%%%%%%%%%%

Table \ref{tab3} lists five data sets together with the experimental
results. These are real data sets on a number of different domains.
\begin{table}[tb] \centering%
\caption{\it\small Incomplete data sets used for the experiments,
together with their number of features, instances, missing values,
and the relative frequency of the majority class. The data sets
are available from the UCI repository of machine learning data
sets \cite{MurAha95}.\label{tab3} Average number of features
selected by the filters on the entire data set are reported in the
last three columns. FF always selected fewer features than F; F
almost always selected fewer features than BF. Prediction
accuracies where significantly different only for the
Hypothyroidloss data set.\label{tab4}}\medskip\small
\begin{tabular*}{\textwidth}{|l@{\extracolsep\fill}rrrr||rrr|}
\hline
Name & \#feat. & \#inst. & \#m.v. & maj.class & FF & F & BF \\ \hline
{Audiology}          & {69} & {226} & {317}  & {0.212} & {64.3} & {68.0} & {68.7} \\
{Crx}                & {15} & {690} & {67}   & {0.555} & {9.7}  & {12.6} & {13.8} \\
{Horse-colic}        & {18} & {368} & {1281} & {0.630} & {11.8} & {16.1} & {17.4} \\
{\bf Hypothyroidloss}& {23} &{3163} & {1980} & {0.952} & {4.3}  & {8.3} & {13.2}\\
{Soybean-large}      & {35} & {683} & {2337} & {0.135} & {34.2} & {35}   & {35}  \\ \hline
\end{tabular*}
\end{table}
The data sets presenting non-nominal features have been
pre-discretized by MLC++ \cite{KoJoLoMaPf94}, default options
(i.e., the common entropy based discretization). This step may
remove some features judging them as irrelevant, so the number of
features in the table refers to the data sets after the possible
discretization. The instances have been randomly sorted before
starting the experiments.

\begin{figure}[tbh]\epsfxsize=86mm
\centerline{\epsfbox{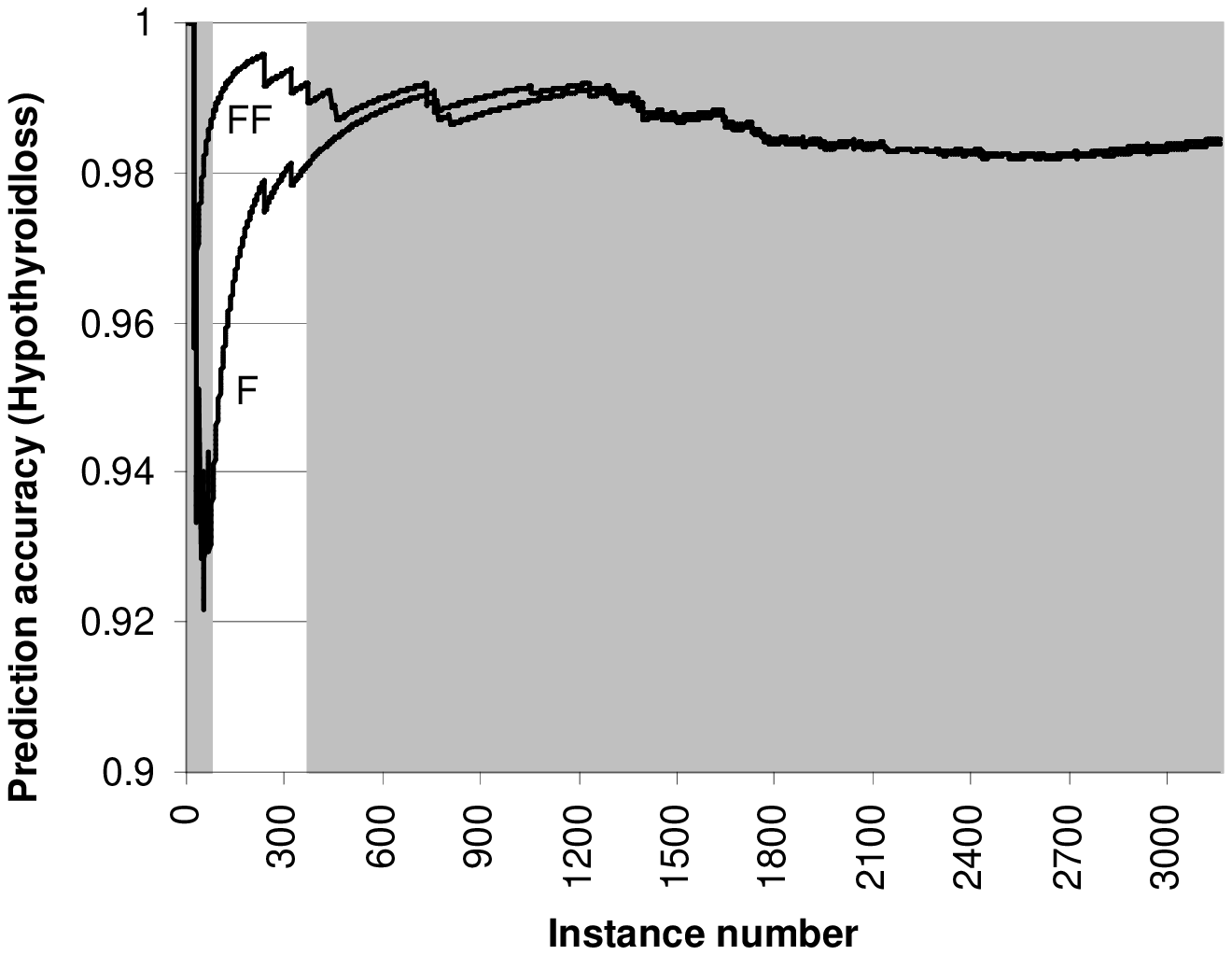}}
\caption{\label{fig5}\it\small Prediction accuracies
of the naive Bayes with filters F and FF on the Hypothyroidloss
data set. BF is not reported because there is no significant
difference with the F curve. The differences between F and FF are
significant in the range of observations 71--374 (white area). The
maximum difference is achieved at observation 71, where the
accuracies are 0.986 (FF)\ vs. 0.930 (F).}
\end{figure}

The last three columns of Table \ref{tab4} show that FF selects
lower (i.e.\ better) number of features than the commonly used filter F,
which in turn, selects lower number of features than the filter
BF. We used the \emph{two-tails paired t test} at level 0.05 to
compare the prediction accuracies of the naive Bayes with
different filters, in the first $k$ instances of the data set, for
each $k$. On four data sets out of five, both the differences
between FF and F, and the differences between F and BF, were never
statistically significant, despite the different number of used
features, as indicated in Table \ref{tab4}. The reduction can be very
pronounced, as for the Hypothyroidloss data set. This is also the
only data set for which the prediction accuracies of F and FF are
significantly different, in favor of the latter. This is displayed
in Figure \ref{fig5}. Similar (even stronger) results have been
found for 10 complete data sets analyzed in
\cite{Hutter:02feature}.

%Remark
The most prominent evidence from the experiments is the better
performance of FF versus the traditional filter F. In the
following we look at FF from another perspective to exemplify and
explain its behavior.
FF includes a feature if $p(I>\varepsilon|\v n)\geq\bar p$,
according to its definition. Let us assume that FF is realized by
means of the Gaussian (as in the experiments above), and let us
choose $\bar p\approx 0.977$. The condition $p(I>\varepsilon|\v
n)\geq\bar p$ becomes $\varepsilon\leq E[I]-2\cdot
\sqrt{\Var[I]}$, or, in an approximate way,
$I(\hat{\v\t})\geq\varepsilon+2\cdot \sqrt{\Var[I]}$, given that
$I(\hat{\v\t})$ is the first-order approximation of $E[I]$ (cf.
\eqref{EI}). We can regard $\varepsilon+2\cdot \sqrt{\Var[I]}$ as
a new threshold $\varepsilon'$. Under this interpretation, we see
that FF is approximately equal to using the filter F with the
bigger threshold $\varepsilon'$. This interpretation makes it also
clearer why FF can be better suited than F for sequential learning
tasks. In sequential learning, $\Var[I]$ decreases as new units
are read; this makes $\varepsilon'$ to be a self-adapting
threshold that adjusts the level of caution (in including
features) as more units are read. In the limit, $\varepsilon'$ is
equal to $\varepsilon$. This characteristic of self-adaptation,
which is absent in F, seems to be decisive to the success of FF.

%%%%%%%%%%%%%%%%%%%%%%%%%%%%%%%%%%%%%%%%%%%%%%%%%%%%%%%%%%%%%%%
\section{Conclusions}\label{secConc}
%%%%%%%%%%%%%%%%%%%%%%%%%%%%%%%%%%%%%%%%%%%%%%%%%%%%%%%%%%%%%%%

We addressed the problem of the reliability of empirical estimates
for the chances $\v\t$ and the mutual information $I$ in the case
of incomplete discrete data. We used the Bayesian framework to
derive reliable and quickly computable approximations for the
mean, mode and the (co)variance of $\v\t$ and $I(\v\t)$ under the
posterior distribution $p(\v\t|D)$. We showed that $p(\v\t|\v D)$
is unimodal, which implies that EM always converges to the global
maximum. The results allowed us to efficiently determine credible
intervals for $I$ with incomplete data. Applications are manifold,
e.g.\ to robustly infer classification trees or Bayesian networks.
As far as feature selection is concerned, we empirically showed
that the forward filter, which includes a feature if the mutual
information is greater than $\eps$ with high probability,
outperforms the popular filter based on empirical mutual
information in sequential learning tasks. This result for
incomplete data is obtained jointly with the naive Bayes
classifier. More broadly speaking, obtaining the distribution of
mutual information when data are incomplete may form a basis on
which reliable and effective uncertain models can be developed.

%%%%%%%%%%%%%%%%%%%%%%%%%%%%%%%%%%%%%%%%%%%%%%%%%%%%%%%%%%%%%%%
%         Bibliography        %
%%%%%%%%%%%%%%%%%%%%%%%%%%%%%%%%%%%%%%%%%%%%%%%%%%%%%%%%%%%%%%%
\begin{small}
%\bibliographystyle{alpha}
%\bibliography{../../hutter,../bayes}
\newcommand{\etalchar}[1]{$^{#1}$}

\end{small}

\end{document}